%% file: main.tex
\pdfoutput=1

\documentclass[11pt]{article}

\usepackage[]{acl}
\usepackage{authblk}
\usepackage{times}
\usepackage{latexsym}
\usepackage{amsmath}
\usepackage{booktabs}
\usepackage{amssymb}
\usepackage{graphicx}
\usepackage[frozencache,cachedir=.]{minted}

\usepackage{multirow}

\newlength{\myMheight}
\newcommand{\github}{\includegraphics[height=\myMheight]{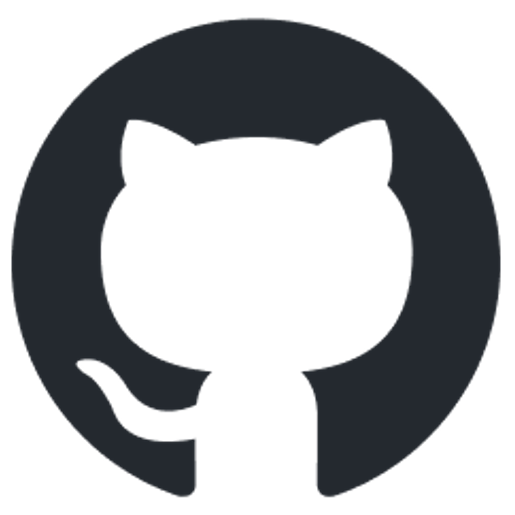}}
\newcommand{\hf}{\includegraphics[height=\myMheight]{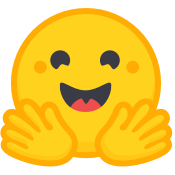}}

\newcommand{\minus}{\scalebox{0.75}[1.0]{$-$}}

\usepackage[T1]{fontenc}
\usepackage[utf8]{inputenc}
\usepackage{microtype}
\newcommand{\inline}[1]{\mintinline{python}{#1}}
\title{
Scaling Synthetic Logical Reasoning Datasets \\
with Context-Sensitive Declarative Grammars}

\author[]{Damien Sileo}
\affil[]{Univ. Lille, Inria, CNRS, Centrale Lille, UMR 9189 - CRIStAL, F-59000 Lille, France}
\affil[]{\url{damien.sileo@inria.fr}}

\begin{document}
\maketitle
\begin{abstract}
Logical reasoning remains a challenge for natural language processing, but it can be improved by training language models to mimic theorem provers on procedurally generated problems. Previous work used domain-specific proof generation algorithms, which biases reasoning toward specific proof traces and limits auditability and extensibility. We present a simpler and more general declarative framework with flexible context-sensitive rules binding multiple languages (specifically, simplified English and the TPTP theorem-proving language). We construct first-order logic problems by selecting up to 32 premises and one hypothesis. We demonstrate that using semantic constraints during generation and careful English verbalization of predicates enhances logical reasoning without hurting natural English tasks.  We use relatively small DeBERTa-v3 models to achieve state-of-the-art accuracy on the FOLIO human-authored logic dataset, surpassing GPT-4 in accuracy with or without an external solver by $12\%$.
\end{abstract}

\section{Introduction}

Language models trained only on natural language show lackluster capabilities at logical reasoning \cite{mccoy2023embers, mahowald2024dissociating}. As a countermeasure, we can train neural models to match the output of symbolic reasoning systems (e.g., logic theorem provers, or other algorithms) on procedurally generated problems, to sharpen their reasoning capabilities. This process improves accuracy on some human-authored problems \cite{wu2021lime,ruletaker,wu2022insights,liu2023zero}. 

Previous work on synthetic first-order logic (FOL) reasoning datasets, RuleTaker \cite{ruletaker}, LogicNLI \cite{goodwin2020probing} and FLD \cite{fld23}, write dedicated code re-implementing the FOL axioms from scratch to generate proofs, and translate the generated problems to natural language. We propose Unigram, a framework for synthetic reasoning data generation, specifically designed to generate problems jointly into multiple languages. We represent grammars with concise and expressive rules binding two languages, and constraints to prune unwanted generations. 






We write the most extensive grammar of FOL semantic fragments to our knowledge.  We structure the generated expressions into \textsc{Premise}, \textsc{Hypothesis} pairs, and annotate their logical relationship (entailment/contradiction/neutral) with a FOL solver, following the natural language inference (NLI) framework \cite{goodwin-etal-2020-probing}. A simplistic FOL NLI problem is:
\textsc{Premise}: \textit{Everyone who is happy is rich. Mary is rich.} \textsc{Hypothesis}: \textit{Mary is happy} \textsc{Label}: \textit{Neutral}.

We fine-tune DeBERTa NLI models \cite{he2021debertav3} on Unigram-FOL and compare it with previous similar datasets. The 184M parameters (base-size) beats GPT-4 augmented or not with external theorem provers, on the FOLIO \cite{han2022folio} dataset. Our contributions are as follows: (i) A dataset of reasoning problems expressed in English and TPTP (a language that can be interfaced with numerous theorem provers) alongside Vampire proof annotations, covering FOL with equality and both finite and open domains, improved compositionality, and more extensive quantifiers. (ii) Ablations measuring the effect of constraining material conditionals usage, of using realistic English predicates, and of reimplementing LogicNLI with declarative generation instead of proof tree generation, highlighting that declarative can work better but that a richer logical modeling drives most of the improvement. (iii) A general reasoning problem grammar-based generation framework relying on solvers.
The generation library, grammars, models, and generated dataset are publicly available\footnote{\href{https://github.com/sileod/unigram}{[code:GitHub \github ]}\href{https://huggingface.co/unigram}{[data:HF-datasets \hf]}\label{footnote:urls}}.

\section{Related work}

\paragraph{Synthetic datasets for reasoning}


Numerous works investigate the logical capabilities of NLP models using textual datasets and symbolic reasoning \cite{helwe2022logitorch}. We focus on the grammar-derived synthetic datasets. RuleTaker \citep{ruletaker} explores this area with a subset of first-order logic. LogicNLI addresses a broader FOL subset \citep{tian-etal-2021-diagnosing}.  FLD explores full FOL \cite{fld23} and increased compositionality. \citet{fragmentsaaai,richardson2022pushing} use a solver to study the satisfiability in natural language using the Z3 solver and dedicated generation logic on constrained problems. Other work explore non-standard logic with synthetic dataset, notably probabilistic \citep{sileo2022probing}, paraconsistant \cite{kazemi2024boardgameqa}, epistemic \cite{sileo-lernould-2023-mindgames} logics.

\paragraph{Generation frameworks}
Multiple frameworks already implement generation from handwritten grammars. NLTK \citep{bird-loper-2004-nltk} has a context-free grammar tool, but cannot natively handle multiple languages or large-scale generation. Grammatical Framework \citep{ranta2004grammatical} is the closest tool to ours. It enables generation from abstract grammars and linearization into concrete grammars (e.g. French and English) but it is translation-oriented and not context-sensitive. GLIF \citep{schaefer2020glif} extends Grammatical Framework to parse English into logical formulas but is not suited for generation either.


\section{Scalable dataset generation without forward inference}

\subsection{Forward inference}
Previous NLI-style FOL reasoning datasets (RuleTaker, LogicNLI, FLD) generate examples using proof generators that are based on the axioms of FOL.
This requires domain-specific generation code and introduces unwanted complexity.  Elimination and Introduction rules can cancel each other and create an illusion of reasoning depth. We found that some examples in the Proofwriter dataset \cite{tafjord-etal-2021-proofwriter} directly contain the premise in the hypothesis despite having a reasoning depth of 5.
When constructing NLI pairs, generating neutral examples requires special strategies introducing a sampling bias, and it can be the same for contradiction generation. Proof generation techniques enable high reasoning depth but at the cost of breadth (linguistic variety and reasoning variety).


\subsection{Declarative generation}
We fully rely on an existing FOL solver and we propose Unigram, a simpler, more generic method to generate problems with multilingual grammars where rules bind multiple surface form realization templates. A Unigram Rule declaration specifies a type signature, and two surface form realizers, and optional validity constraints:
$\textbf{R}(\text{output\_type, input\_types, realizers, constraints})$
The signature specifies the type of the rule output and the type of the arguments. The realizers take the arguments as input and map them to a string. We can have a realizer for a logic code and a realizer for English. 
Using functions allows more expressivity than context-free grammars \citep{hunter2021chomsky}, but for most cases with can treat template strings as functions using Python \inline{string.format}.
Constraints and realizers can access the state of the current generation as an \href{https://anytree.readthedocs.io/}{anytree} tree. Constraints are binary functions checking construction validity. One useful constraint is distinctness, e.g. (arguments of the same type should have a different realization), to avoid repetitions or statements like \textit{Mary likes Mary}. We enable this constraint by default.



\paragraph{Generation algorithm}
We use a depth-first algorithm that recursively fills in the leftmost non-terminal leave with random type-matching rule sampling until constraints are satisfied. This enables left-to-right generation, allowing realizers and constraints to access the current context. We recursively call realizers to construct surface forms (e.g. English text).




\section{Application to first-order logic (FOL)}

We use Unigram to enrich FOL problem generation while also avoiding ambiguity, starting as a superset of LogicNLI (grammar in Appendix \ref{sec:lnli}). To create a problem, we uniformly sample 1 to 32 sentences as premises and 1 sentence per hypothesis ensuring that all symbols are present in the premise. We exclude non-satisfiable formulas (paradoxes) in premise groups and hypotheses. We label pairs as \textsc{entailment} if $(premise \land \neg hypothesis)$ is unsatisfiable, as \textsc{contradiction} if
$(premise \land hypothesis)$ is unsatisfiable, and as \textsc{neutral} otherwise.
Following Ruletaker and LogicNLI, we create problems with predicates over named individuals (e.g. \textit{Mary is young}). We generate gender-balanced English surnames with \href{https://pypi.org/project/censusname/}{CensusName}.
We now present new logical modeling features absent from the previous comparable datasets:

\paragraph{Explicit finite and open domains}

We explicitly mention the domain when using the quantifiers. We introduce two locations, \textit{anywhere}, and a \textit{room} with occupants e.g. \textit{Mary, and Paul are the only persons in the room.} which logically means \(\forall x, room(x) \to (x = \text{Mary} \lor x = \text{Paul}) \). We can then quantify over the room (\textit{everyone in the room}) or anywhere (\textit{everyone anywhere}). By doing this, we can generate induction problems (checking that \text{everyone in the room is happy} if \text{Mary and Paul are happy}) and test reasoning with both finite and open domains. This requires handling FOL with equality which was not implemented in previous work.

\paragraph{Quantifiers and logical relationships}
We extend previous work with more complete quantifiers \textit{not all, nobody, not everyone}.  We leverage context-sensitivity to create a rule for polysyllogisms (predicate chains of the form \text{all A are B, all B are C, all C are D}. We also introduce \textit{only if, unless, otherwise} as conditionals and allow sentence-level negation.

\paragraph{Constraining material conditionals}
Like previous work, we use material conditional to express conditional statements: \textit{if p then q} is formalized as $p\to q$ i.e.  $\neg p \lor q$. This means that the implication is true if $p$ is false, and that negating  $p\to q$ entails $q$ both $\neg p$ which can be counter-intuitive. We use a constraint to eliminate all conditionals within the scope of negations and of other conditionals.

\paragraph{Improving predicate verbalization}
RuleTaker and LogicNLI use adjectives as logical predicates but do not handle their semantic interference. RuleTaker do not consider being both \textit{blue} and being \textit{green} as contradictory. LogicNLI uses 379 adjectives treated as independent, including \textit{ugly} and \textit{ugliest}. FLD uses pseudo language like \textit{	
the lard does hurtle pushup}. We prompted GPT-4 (May version) to \textit{Generate 150 predicates where each predicate does not contradict nor entail any other predicate. Two examples: "enjoys wildlife photography" and "owns a smart tv".} We remove errors and provide manual negations. We also use relationships (\textit{like, is a sibling of}, modeling symmetry axioms when relevant, and adjectives.

\paragraph{Logical representation language}
Previous LogicNLI, RuleTaker, FLD, and FOLIO all use their own logical format, representing formulas as lists or strings. We use the TPTP \cite{sutcliffe2010tptp} TPTP FOF language which is a standard syntax for theorem provers evaluation and is compatible with many theorem provers, notably Vampire \cite{reger2022vampire}, Z3 \cite{de2008z3} or Prover9 \cite{mccune2005release}. We select the Vampire \cite{reger2022vampire} theorem prover which provides short and readable proofs and details all the premises used during a derivation.

\paragraph{Complexity control}

Methods based on forward inference can theoretically control the proof depth using hyperparameters. Here, to avoid mostly sampling shallow problems, we limit the number of non-neutral examples where the proof to the number of examples using 5 inputs, for each number of inputs. Neutrals are still a majority by an order of magnitude. To sample hard neutral examples, we use a Gradient Boosting classifier with 100 trees (and scikit-learn 1.5.0 \cite{pedregosa2011scikit} default parameters otherwise) to predict the labels based on unigram counts of the logical operators in the premise and hypothesis. We train on $1k$ examples, discard these, and then discard the most confident neutral predictions to achieve balanced labels.

\section{Experiments}
\begin{footnotesize}
\begin{figure*}
    \centering
    \begin{bitsmall}
    \include{tab}
    \end{bitsmall}
    \vspace{-5mm}
    \caption{Comparison of auxiliary synthetic training datasets effect on the evaluation tasks. We report the average accuracy of two runs. $\mathcal{D}$ column refer to zero-shot $\mathcal{D}$ test accuracy after synthetic auxiliary training, and +ft refers to the test accuracy after auxiliary training then further fine-tuning $\mathcal{D}$ training set (in the previous column).
    }    \vspace{-1mm}

    \label{fig:results}
\end{figure*}
\end{footnotesize}

\subsection{Methodology}

We fine-tune a pre-trained NLI model on multiple synthetic FOL datasets: LogicNLI, FLD, RuleTaker, and on Unigram-FOL. We then evaluate the direct effect on other three-way entailment downstream tasks, and on further fine-tuning on the training data of evaluation tasks \cite{phang2018sentence}.

We use the DeBERTa-v3 \cite{he2021debertav3} NLI models trained on the tasksource collection \citep{sileo2024tasksource}\footnote{
\href{https://hf.co/sileod/deberta-v3-base-tasksource-nli}{hf.co/deberta-v3-base-tasksource-nli}}.  We use a learning rate of $1e{-}5$ for DeBERTa-large and $2e{-}5$ \cite{mosbach2021on} for DeBERTa-base, 1 or 3 epochs (based on intrinsic validation accuracy) and Huggingface Transformers \cite{wolf2019huggingface} version 4.41 \href{https://github.com/huggingface/transformers/blob/dc6eb448419b6a4a4967cdc7e8b96bdbc6969644/src/transformers/training_args.py#L850}{default} Trainer arguments otherwise.


We generate $100k$ examples with a $80/10/10$ train/dev/test split.
but we only use $40k$ training training examples to match FLD. We use the FLD$\star$ version of FLD.
We use the ProofWriter \cite{tafjord-etal-2021-proofwriter} open-world-assumption version of RuleTaker. We exclude LogicNLI examples labeled as paradoxes and we map all labels to NLI labels.

\subsection{Evaluation datasets}
\vspace{-1mm}
We evaluate on two pure reasoning datasets, FOLIO and Fragments, and on two more general datasets:
\textbf{FOLIO} \citep{han2022folio} contains human-written FOL problems. We evaluate on the validation set to compare to \citet{olausson-etal-2023-linc} results who report $72.5\%$ accuracy using a GPT-4 with a solver and $75.3\%$ with chain-of-thoughts. We construct another validation set from $10\%$ of train and map labels to NLI labels. \cite{wei2022chain} \textbf{WANLI} \cite{liu-etal-2022-wanli} is a NLI dataset with diverse and challenging reasoning patterns. \textbf{ConTRoL} \cite{Liu_Cui_Liu_Zhang_2021} is a NLI dataset requiring multiple premises to derive the correct label, measuring contextual reasoning. \textbf{Fragments} \cite{fragmentsaaai} is based on formal semantics templates and evaluate reasoning with quantifiers; this dataset is mostly suited to evaluation, as training quickly leads to almost perfect test accuracy.

\vspace{-1mm}
\paragraph{Comparison with previous synthetic datasets} Table \ref{fig:results} shows the accuracy of multiple auxiliary training datasets on the evaluation dataset.
Unigram-FOL outperforms RuleTaker, LogicNLI, and FLD on all tasks with a comfortable margin, and leads to lesser degradation on the datasets that are not only focused on logic (WANLI, ConTRoL). The last line of the table combines Unigram-FOL (with the full 100k examples) with FLD and shows that combining generation methods can further push the state of the art on FOLIO.

We conduct ablations to better understand the source of this improvement, presented in the middle of Table \ref{fig:results}.\vspace{-2mm}

\paragraph{Unigram-LogicNLI} We use our declarative generation method on the base LogicNLI grammar to disentangle the effect of the generation technique from the grammar itself. This outperforms the original LogicNLI but not Unigram-FOL which highlights the value of our additional constructions.\vspace{-1mm}
\paragraph{Replacing Realistic Predicates} We replace our generated predicates with the original LogicNLI adjectives (containing semantic interferences); this degrades FOLIO accuracy but does not strongly impact other NLI tasks, notably Fragments which mainly use adjectives as predicates.\vspace{-2mm}
\paragraph{Removing Conditionals Constraints} Unrestricting usage of material conditionals harms the zero-shot transfer on FOLIO and the capabilities at more general reasoning, which confirms that removing counter-intuitive constructs can help transferability.
\vspace{-3mm}
\section{Conclusion}
\vspace{-3mm}
We showed that simple declarative grammars paired with solvers can outperform complex proof tree generators for reasoning dataset generations and released a new FOL reasoning dataset, models, and ablations. Our framework can help future reasoning research, notably on explanation since fully aligned TPTP code can be leveraged to model necessity and sufficiency. We plan to extend Unigram to planning, constraint satisfaction and modal logic. 





\newpage
\section*{Limitations}

Reasoning methods based on neural networks do not provide formal guarantees and can introduce biases in real applications. They can be used as a complement to externalization methods \cite{olausson-etal-2023-linc}. Automatically formalizing a problem is difficult and can lead to mistakes \cite{olausson-etal-2023-linc} which could be detected by internalization-based methods. Our dataset could be used to automate formalization but we did not try such experiments. In addition, our work is only conducted with English language and encoder models, mainly used for verification and not generation. We only used one model architecture, DeBERTa, while other architectures like Albert \cite{Lan2020ALBERT:} or other recursive architectures could be more suited to reasoning.

\section*{Ethical considerations}
Our models are derived from language models which inherit bias from their training corpus. We did not conduct any human annotations, relying on already annotated datasets to validate our methodology. We use encoder models which have lower energy consumption than decoders \cite{luccioni2024power} and performed experiments with less than 20 total days on a Nvidia A100 GPU.

\bibliography{anthology,custom}

\onecolumn \appendix

\section{Unigram LogicNLI reimplementation \label{sec:lnli}}
\input{lnli}



\end{document}

%% file: tab.tex
\begin{tabular}{lllllllll}
\toprule
Model size & Auxilary training & FOLIO & +ft & WANLI & +ft & CTRL & +ft & Fragments \\
\midrule
D-base & - & 49.5 & \underline{74.3} & \textbf{65.2} & \textbf{77.4} & \textbf{46.2} & \textbf{56.7} & \underline{63.6} \\
D-base & RuleTaker & 55.1 & 71.3 & 60.9 & 73.8 & 36.0 & 53.0 & 48.7 \\
D-base & LogicNLI & 50.5 & 69.3 & 61.1 & 72.4 & 38.4 & 54.4 & 56.3 \\
D-base & FLD & \underline{59.9} & 72.3 & 60.0 & 73.6 & 38.2 & 55.8 & 56.8 \\
D-base & Unigram-FOL & \textbf{64.4} & \textbf{78.2} & \underline{63.6} & \underline{75.6} & \underline{42.8} & \underline{56.6} & \textbf{65.4} \\
\midrule
D-base & $\minus$ Constrained\_Conditionals & 63.4 & 81.2 & 62.2 & 71.8 & 40.6 & 55.4 & 59.8 \\
D-base & $\minus$ Realistic\_Predicates & 62.4 & 76.2 & 65.8 & 74.4 & 41.8 & 53.2 & 68.2 \\
D-base & Unigram-LogicNLI & 57.4 & 71.3 & 61.6 & 76.4 & 38.6 & 55.6 & 57.8 \\
\midrule
D-large & - & 49.5 & 70.0 & 66.2 & 77.0 & \textbf{49.6} & \underline{62.0} & \underline{67.6} \\
D-large & RuleTaker & 58.1 & 77.2 & 68.5 & \underline{77.9} & 43.1 & 60.7 & 61.7 \\
D-large & LogicNLI & 58.7 & 73.3 & \underline{68.5} & 77.4 & 45.4 & 60.9 & 64.4 \\
D-large & FLD & \underline{60.9} & \underline{78.2} & 68.0 & 77.6 & 44.0 & 59.8 & 61.7 \\
D-large & Unigram-FOL & \textbf{63.4} & \textbf{82.2} & \textbf{75.4} & \textbf{81.6} & \underline{48.2} & \textbf{62.2} & \textbf{73.2} \\
\midrule
D-large & Unigram-FOL+FLD & 78.2 & 88.6 & 65.2 & 78.4 & 42.2 & 57.9 & 75.4 \\
\bottomrule
\end{tabular}

%

%% file: lnli.tex
\begin{minted}{python}

from unigram import Rule as R

ADJECTIVES = ['rich','quiet','old','tall','kind','brave','wise',
'happy', 'strong','curious','patient','funny','generous','humble']
# (We selected adjectives with no clear semantic interference)
NAMES = ['mary', 'paul', 'fred', 'alice', 'john', 'susan', 'lucy']

R.init(['tptp','eng'], "fof") 

R('premise(' + ','.join(['rule']*16')'+','+'.'join(['fact]*8)+')',
  '&\n'.join([f'({i})' for i in range(24)]),
   '\n'.join([f'{i}' for i in range(24)]))

R('hypothesis(person,a)','1(0)','0 is 1')

for a in ADJECTIVES:
    R('adj', a), R('adj', f'~{a}', f'not {a}', weight=0.2)
    
R('property(adj,adj)', '(0(?)&1(?))', 'both 0 and 1')
R('property(adj,adj)', '(0(?)|1(?))', '0 or 1')
R('property(adj,adj)', '(0(?)<~>1(?))', 'either 0 or 1', weight=0.5)
R('property(adj)', '0(?)', '0')

R('rule(property,property)', '![X]:(0[?←X]=>1[?←X])',
'everyone who is 0 is 1')
R('rule(property,property)', '![X]:(0[?←X]<=>1[?←X])',
'everyone who is 0 is 1 and vice versa')

for p in NAMES:
    R('person', p)
    
R('fact(person,property)','1[?←0]', '0 is 1')
R('fact(property)', '?[X]:(0[?←X])', 'someone is 0', weight=0.2)

R('rule(fact,fact)', '(0)=>(1)', 'if 0 then 1')
R('rule(fact,fact)', '(0)<=>(1)', 'if 0 then 1 and vice versa')
\end{minted}

%% file: main.bbl
\begin{thebibliography}{38}
\expandafter\ifx\csname natexlab\endcsname\relax\def\natexlab#1{#1}\fi

\bibitem[{Bird and Loper(2004)}]{bird-loper-2004-nltk}
Steven Bird and Edward Loper. 2004.
\newblock \href {https://aclanthology.org/P04-3031} {{NLTK}: The natural language toolkit}.
\newblock In \emph{Proceedings of the {ACL} Interactive Poster and Demonstration Sessions}, pages 214--217, Barcelona, Spain. Association for Computational Linguistics.

\bibitem[{Clark et~al.(2020)Clark, Tafjord, and Richardson}]{ruletaker}
Peter Clark, Oyvind Tafjord, and Kyle Richardson. 2020.
\newblock \href {https://doi.org/10.24963/ijcai.2020/537} {Transformers as soft reasoners over language}.
\newblock In \emph{Proceedings of the Twenty-Ninth International Joint Conference on Artificial Intelligence, {IJCAI-20}}, pages 3882--3890. International Joint Conferences on Artificial Intelligence Organization.
\newblock Main track.

\bibitem[{De~Moura and Bj{\o}rner(2008)}]{de2008z3}
Leonardo De~Moura and Nikolaj Bj{\o}rner. 2008.
\newblock Z3: An efficient smt solver.
\newblock In \emph{International conference on Tools and Algorithms for the Construction and Analysis of Systems}, pages 337--340. Springer.

\bibitem[{Goodwin et~al.(2020{\natexlab{a}})Goodwin, Sinha, and O'Donnell}]{goodwin2020probing}
Emily Goodwin, Koustuv Sinha, and Timothy~J O'Donnell. 2020{\natexlab{a}}.
\newblock Probing linguistic systematicity.
\newblock \emph{arXiv preprint arXiv:2005.04315}.

\bibitem[{Goodwin et~al.(2020{\natexlab{b}})Goodwin, Sinha, and O{'}Donnell}]{goodwin-etal-2020-probing}
Emily Goodwin, Koustuv Sinha, and Timothy~J. O{'}Donnell. 2020{\natexlab{b}}.
\newblock \href {https://doi.org/10.18653/v1/2020.acl-main.177} {Probing linguistic systematicity}.
\newblock In \emph{Proceedings of the 58th Annual Meeting of the Association for Computational Linguistics}, pages 1958--1969, Online. Association for Computational Linguistics.

\bibitem[{Han et~al.(2022)Han, Schoelkopf, Zhao, Qi, Riddell, Benson, Sun, Zubova, Qiao, Burtell et~al.}]{han2022folio}
Simeng Han, Hailey Schoelkopf, Yilun Zhao, Zhenting Qi, Martin Riddell, Luke Benson, Lucy Sun, Ekaterina Zubova, Yujie Qiao, Matthew Burtell, et~al. 2022.
\newblock Folio: Natural language reasoning with first-order logic.
\newblock \emph{arXiv preprint arXiv:2209.00840}.

\bibitem[{He et~al.(2021)He, Gao, and Chen}]{he2021debertav3}
Pengcheng He, Jianfeng Gao, and Weizhu Chen. 2021.
\newblock Debertav3: Improving deberta using electra-style pre-training with gradient-disentangled embedding sharing.
\newblock \emph{arXiv preprint arXiv:2111.09543}.

\bibitem[{Helwe et~al.(2022)Helwe, Clavel, and Suchanek}]{helwe2022logitorch}
Chadi Helwe, Chlo{\'e} Clavel, and Fabian Suchanek. 2022.
\newblock Logitorch: A pytorch-based library for logical reasoning on natural language.
\newblock In \emph{The 2022 Conference on Empirical Methods in Natural Language Processing: System Demonstrations}.

\bibitem[{Hunter(2021)}]{hunter2021chomsky}
Tim Hunter. 2021.
\newblock The chomsky hierarchy.
\newblock \emph{A companion to Chomsky}, pages 74--95.

\bibitem[{Kazemi et~al.(2024)Kazemi, Yuan, Bhatia, Kim, Xu, Imbrasaite, and Ramachandran}]{kazemi2024boardgameqa}
Mehran Kazemi, Quan Yuan, Deepti Bhatia, Najoung Kim, Xin Xu, Vaiva Imbrasaite, and Deepak Ramachandran. 2024.
\newblock Boardgameqa: A dataset for natural language reasoning with contradictory information.
\newblock \emph{Advances in Neural Information Processing Systems}, 36.

\bibitem[{Lan et~al.(2020)Lan, Chen, Goodman, Gimpel, Sharma, and Soricut}]{Lan2020ALBERT:}
Zhenzhong Lan, Mingda Chen, Sebastian Goodman, Kevin Gimpel, Piyush Sharma, and Radu Soricut. 2020.
\newblock \href {https://openreview.net/forum?id=H1eA7AEtvS} {Albert: A lite bert for self-supervised learning of language representations}.
\newblock In \emph{International Conference on Learning Representations}.

\bibitem[{Liu et~al.(2022)Liu, Swayamdipta, Smith, and Choi}]{liu-etal-2022-wanli}
Alisa Liu, Swabha Swayamdipta, Noah~A. Smith, and Yejin Choi. 2022.
\newblock \href {https://doi.org/10.18653/v1/2022.findings-emnlp.508} {{WANLI}: Worker and {AI} collaboration for natural language inference dataset creation}.
\newblock In \emph{Findings of the Association for Computational Linguistics: EMNLP 2022}, pages 6826--6847, Abu Dhabi, United Arab Emirates. Association for Computational Linguistics.

\bibitem[{Liu et~al.(2021)Liu, Cui, Liu, and Zhang}]{Liu_Cui_Liu_Zhang_2021}
Hanmeng Liu, Leyang Cui, Jian Liu, and Yue Zhang. 2021.
\newblock \href {https://doi.org/10.1609/aaai.v35i15.17580} {Natural language inference in context - investigating contextual reasoning over long texts}.
\newblock \emph{Proceedings of the AAAI Conference on Artificial Intelligence}, 35(15):13388--13396.

\bibitem[{Liu et~al.(2023)Liu, Zhou, Jiang, Dou, and Lin}]{liu2023zero}
Qian Liu, Fan Zhou, Zhengbao Jiang, Longxu Dou, and Min Lin. 2023.
\newblock \href {https://arxiv.org/abs/2304.07995} {From zero to hero: Examining the power of symbolic tasks in instruction tuning}.
\newblock \emph{arXiv preprint arXiv:2304.07995}.

\bibitem[{Luccioni et~al.(2024)Luccioni, Jernite, and Strubell}]{luccioni2024power}
Sasha Luccioni, Yacine Jernite, and Emma Strubell. 2024.
\newblock Power hungry processing: Watts driving the cost of ai deployment?
\newblock In \emph{The 2024 ACM Conference on Fairness, Accountability, and Transparency}, pages 85--99.

\bibitem[{Mahowald et~al.(2024)Mahowald, Ivanova, Blank, Kanwisher, Tenenbaum, and Fedorenko}]{mahowald2024dissociating}
Kyle Mahowald, Anna~A Ivanova, Idan~A Blank, Nancy Kanwisher, Joshua~B Tenenbaum, and Evelina Fedorenko. 2024.
\newblock Dissociating language and thought in large language models.
\newblock \emph{Trends in Cognitive Sciences}.

\bibitem[{McCoy et~al.(2023)McCoy, Yao, Friedman, Hardy, and Griffiths}]{mccoy2023embers}
R~Thomas McCoy, Shunyu Yao, Dan Friedman, Matthew Hardy, and Thomas~L Griffiths. 2023.
\newblock Embers of autoregression: Understanding large language models through the problem they are trained to solve.
\newblock \emph{arXiv preprint arXiv:2309.13638}.

\bibitem[{McCune(2005)}]{mccune2005release}
William McCune. 2005.
\newblock Release of prover9.
\newblock In \emph{Mile high conference on quasigroups, loops and nonassociative systems, Denver, Colorado}.

\bibitem[{Morishita et~al.(2023)Morishita, Morio, Yamaguchi, and Sogawa}]{fld23}
Terufumi Morishita, Gaku Morio, Atsuki Yamaguchi, and Yasuhiro Sogawa. 2023.
\newblock Learning deductive reasoning from synthetic corpus based on formal logic.
\newblock In \emph{International Conference on Machine Learning}, pages 25254--25274. PMLR.

\bibitem[{Mosbach et~al.(2021)Mosbach, Andriushchenko, and Klakow}]{mosbach2021on}
Marius Mosbach, Maksym Andriushchenko, and Dietrich Klakow. 2021.
\newblock \href {https://openreview.net/forum?id=nzpLWnVAyah} {On the stability of fine-tuning {\{}bert{\}}: Misconceptions, explanations, and strong baselines}.
\newblock In \emph{International Conference on Learning Representations}.

\bibitem[{Olausson et~al.(2023)Olausson, Gu, Lipkin, Zhang, Solar-Lezama, Tenenbaum, and Levy}]{olausson-etal-2023-linc}
Theo Olausson, Alex Gu, Ben Lipkin, Cedegao Zhang, Armando Solar-Lezama, Joshua Tenenbaum, and Roger Levy. 2023.
\newblock \href {https://doi.org/10.18653/v1/2023.emnlp-main.313} {{LINC}: A neurosymbolic approach for logical reasoning by combining language models with first-order logic provers}.
\newblock In \emph{Proceedings of the 2023 Conference on Empirical Methods in Natural Language Processing}, pages 5153--5176, Singapore. Association for Computational Linguistics.

\bibitem[{Pedregosa et~al.(2011)Pedregosa, Varoquaux, Gramfort, Michel, Thirion, Grisel, Blondel, Prettenhofer, Weiss, Dubourg et~al.}]{pedregosa2011scikit}
Fabian Pedregosa, Ga{\"e}l Varoquaux, Alexandre Gramfort, Vincent Michel, Bertrand Thirion, Olivier Grisel, Mathieu Blondel, Peter Prettenhofer, Ron Weiss, Vincent Dubourg, et~al. 2011.
\newblock Scikit-learn: Machine learning in python.
\newblock \emph{the Journal of machine Learning research}, 12:2825--2830.

\bibitem[{Phang et~al.(2018)Phang, F{\'e}vry, and Bowman}]{phang2018sentence}
Jason Phang, Thibault F{\'e}vry, and Samuel~R Bowman. 2018.
\newblock Sentence encoders on stilts: Supplementary training on intermediate labeled-data tasks.
\newblock \emph{arXiv preprint arXiv:1811.01088}.

\bibitem[{Ranta(2004)}]{ranta2004grammatical}
Aarne Ranta. 2004.
\newblock Grammatical framework.
\newblock \emph{Journal of Functional Programming}, 14(2):145--189.

\bibitem[{Reger et~al.(2022)Reger, Suda, Voronkov, Kov{\'a}cs, Bhayat, Gleiss, Hajdu, Hozzova, Evgeny~Kotelnikov, Rawson et~al.}]{reger2022vampire}
Giles Reger, Martin Suda, Andrei Voronkov, Laura Kov{\'a}cs, Ahmed Bhayat, Bernhard Gleiss, Marton Hajdu, Petra Hozzova, JR~Evgeny~Kotelnikov, Michael Rawson, et~al. 2022.
\newblock Vampire 4.7-smt system description.

\bibitem[{Richardson et~al.(2020)Richardson, Hu, Moss, and Sabharwal}]{fragmentsaaai}
Kyle Richardson, Hai Hu, Lawrence Moss, and Ashish Sabharwal. 2020.
\newblock \href {https://doi.org/10.1609/aaai.v34i05.6397} {Probing natural language inference models through semantic fragments}.
\newblock \emph{Proceedings of the AAAI Conference on Artificial Intelligence}, 34:8713--8721.

\bibitem[{Richardson and Sabharwal(2022)}]{richardson2022pushing}
Kyle Richardson and Ashish Sabharwal. 2022.
\newblock Pushing the limits of rule reasoning in transformers through natural language satisfiability.
\newblock In \emph{Proceedings of the AAAI Conference on Artificial Intelligence}, volume~36, pages 11209--11219.

\bibitem[{Schaefer and Kohlhase(2020)}]{schaefer2020glif}
Jan~Frederik Schaefer and Michael Kohlhase. 2020.
\newblock Glif: A declarative framework for symbolic natural language understanding.
\newblock In \emph{FCR@ KI}, pages 4--11.

\bibitem[{Sileo(2024)}]{sileo2024tasksource}
Damien Sileo. 2024.
\newblock tasksource: A large collection of nlp tasks with a structured dataset preprocessing framework.
\newblock In \emph{Proceedings of the 2024 Joint International Conference on Computational Linguistics, Language Resources and Evaluation (LREC-COLING 2024)}, pages 15655--15684.

\bibitem[{Sileo and Lernould(2023)}]{sileo-lernould-2023-mindgames}
Damien Sileo and Antoine Lernould. 2023.
\newblock \href {https://doi.org/10.18653/v1/2023.findings-emnlp.303} {{M}ind{G}ames: Targeting theory of mind in large language models with dynamic epistemic modal logic}.
\newblock In \emph{Findings of the Association for Computational Linguistics: EMNLP 2023}, pages 4570--4577, Singapore. Association for Computational Linguistics.

\bibitem[{Sileo and Moens(2023)}]{sileo2022probing}
Damien Sileo and Marie-francine Moens. 2023.
\newblock \href {https://doi.org/10.18653/v1/2023.starsem-1.41} {Probing neural language models for understanding of words of estimative probability}.
\newblock In \emph{Proceedings of the 12th Joint Conference on Lexical and Computational Semantics (*SEM 2023)}, pages 469--476, Toronto, Canada. Association for Computational Linguistics.

\bibitem[{Sutcliffe(2010)}]{sutcliffe2010tptp}
Geoff Sutcliffe. 2010.
\newblock The tptp world--infrastructure for automated reasoning.
\newblock In \emph{International Conference on Logic for Programming Artificial Intelligence and Reasoning}, pages 1--12. Springer.

\bibitem[{Tafjord et~al.(2021)Tafjord, Dalvi, and Clark}]{tafjord-etal-2021-proofwriter}
Oyvind Tafjord, Bhavana Dalvi, and Peter Clark. 2021.
\newblock \href {https://doi.org/10.18653/v1/2021.findings-acl.317} {{P}roof{W}riter: Generating implications, proofs, and abductive statements over natural language}.
\newblock In \emph{Findings of the Association for Computational Linguistics: ACL-IJCNLP 2021}, pages 3621--3634, Online. Association for Computational Linguistics.

\bibitem[{Tian et~al.(2021)Tian, Li, Chen, Xiao, He, and Jin}]{tian-etal-2021-diagnosing}
Jidong Tian, Yitian Li, Wenqing Chen, Liqiang Xiao, Hao He, and Yaohui Jin. 2021.
\newblock \href {https://doi.org/10.18653/v1/2021.emnlp-main.303} {Diagnosing the first-order logical reasoning ability through {L}ogic{NLI}}.
\newblock In \emph{Proceedings of the 2021 Conference on Empirical Methods in Natural Language Processing}, pages 3738--3747, Online and Punta Cana, Dominican Republic. Association for Computational Linguistics.

\bibitem[{Wei et~al.(2022)Wei, Wang, Schuurmans, Bosma, Xia, Chi, Le, Zhou et~al.}]{wei2022chain}
Jason Wei, Xuezhi Wang, Dale Schuurmans, Maarten Bosma, Fei Xia, Ed~Chi, Quoc~V Le, Denny Zhou, et~al. 2022.
\newblock Chain-of-thought prompting elicits reasoning in large language models.
\newblock \emph{Advances in neural information processing systems}, 35:24824--24837.

\bibitem[{Wolf et~al.(2019)Wolf, Debut, Sanh, Chaumond, Delangue, Moi, Cistac, Rault, Louf, Funtowicz et~al.}]{wolf2019huggingface}
Thomas Wolf, Lysandre Debut, Victor Sanh, Julien Chaumond, Clement Delangue, Anthony Moi, Pierric Cistac, Tim Rault, R{\'e}mi Louf, Morgan Funtowicz, et~al. 2019.
\newblock Huggingface's transformers: State-of-the-art natural language processing.
\newblock \emph{arXiv preprint arXiv:1910.03771}.

\bibitem[{Wu et~al.(2022)Wu, Li, and Liang}]{wu2022insights}
Yuhuai Wu, Felix Li, and Percy~S Liang. 2022.
\newblock \href {https://arxiv.org/abs/2206.10139} {Insights into pre-training via simpler synthetic tasks}.
\newblock \emph{NEURIPS 2022}, 35:21844--21857.

\bibitem[{Wu et~al.(2021)Wu, Rabe, Li, Ba, Grosse, and Szegedy}]{wu2021lime}
Yuhuai Wu, Markus~N Rabe, Wenda Li, Jimmy Ba, Roger~B Grosse, and Christian Szegedy. 2021.
\newblock \href {https://arxiv.org/abs/2101.06223} {Lime: Learning inductive bias for primitives of mathematical reasoning}.
\newblock In \emph{ICML}, pages 11251--11262. PMLR.

\end{thebibliography}
